\documentclass{article}
\usepackage{tcolorbox}

     \usepackage[final]{tackling_climate_workshop_style}

\usepackage[utf8]{inputenc} 
\usepackage[T1]{fontenc}    
\usepackage{hyperref}       
\usepackage{url}            
\usepackage{booktabs}       
\usepackage{amsfonts}       
\usepackage{amsmath}        
\usepackage{nicefrac}       
\usepackage{microtype}      
\usepackage{multirow} 
\usepackage{soul}
\usepackage{graphicx}
\usepackage{longtable}
\title{Geospatial Chain of Thought Reasoning for Enhanced Visual Question Answering on Satellite Imagery}

\author{%
  Shambhavi Shanker\thanks{Work done during an internship at IBM Research India. Correspondence to: Shambhavi Shanker <21d070066@iitb.ac.in>.} \\
  Indian Institute of Technology Bombay \\
  Mumbai, India \\
  \texttt{21d070066@iitb.ac.in}
  \AND
  Manikandan Padmanaban \\
  IBM Research \\
  Bangalore, India \\
  \texttt{manipadm@in.ibm.com}
  \And
  Jagabondhu Hazra \\
  IBM Research \\
  Bangalore, India \\
  \texttt{jahazra1@in.ibm.com}
}

\begin{document}

\maketitle

\begin{abstract}

Geospatial chain of thought (CoT) reasoning is essential for advancing Visual Question Answering (VQA) on satellite imagery, particularly in climate related applications such as disaster monitoring, infrastructure risk assessment, urban resilience planning, and policy support. Existing VQA models enable scalable interpretation of remote sensing data but often lack the structured reasoning required for complex geospatial queries. We propose a VQA framework that integrates CoT reasoning with Direct Preference Optimization (DPO) to improve interpretability, robustness, and accuracy. By generating intermediate rationales, the model better handles tasks involving detection, classification, spatial relations, and comparative analysis, which are critical for reliable decision support in high stakes climate domains. Experiments show that CoT supervision improves accuracy by 34.9\% over direct baselines, while DPO yields additional gains in accuracy and reasoning quality. The resulting system advances VQA for multispectral Earth observation by enabling richer geospatial reasoning and more effective climate use cases.

\end{abstract}

\section{Introduction}

The growing impacts of climate change, including floods, wildfires, droughts, and extreme weather, demand accurate interpretation of Earth observation data. Satellite imagery provides rich multispectral and temporal views critical for disaster monitoring, risk assessment, and climate resilience, but manual analysis is resource intensive and traditional machine learning pipelines remain narrow and task specific. Vision language models (VLMs) address this gap by enabling natural language queries on imagery with grounded responses, making climate information more accessible and actionable. This is especially vital for domains such as flood mapping \cite{floodnet}, wildfire monitoring \cite{wildfire}, and climate adaptation planning \cite{rolnick2022ml}, where timely insights support disaster response and resilience.

Recent advances in multimodal learning have accelerated the integration of VLMs into Earth observation. Models such as EarthDial \cite{soni2024earthdial} highlight this progress by enabling dialogue over remote sensing data and demonstrating strong results in classification, detection, captioning, change detection, and visual question answering. Other efforts, including RemoteCLIP \cite{remoteclip}, RS5M \cite{rs5m}, and GeoChat \cite{geochat}, contribute multimodal datasets and benchmarks for geospatial tasks. Collectively, these developments mark a shift from task-specific pipelines toward general-purpose frameworks capable of handling diverse geospatial queries. Such capabilities are especially relevant for climate applications, where decision makers must synthesize heterogeneous information under uncertainty. For instance, flood response requires reasoning about water, infrastructure, and settlements, while wildfire monitoring demands temporal comparisons of vegetation indices to assess burn severity and risks.

Despite this progress, existing VLMs often lack explicit reasoning. Trained primarily to provide direct answers, they struggle with multi-step inference, causal reasoning, or comparative analysis. In climate related decision making, where outputs must be accurate and trustworthy, this poses risks. Prior work shows that chain-of-thought reasoning improves interpretability and robustness \cite{wang2023selfconsistency, wei2022cot, kojima2022large}, while reinforcement learning methods such as DPO \cite{rafailov2023dpo} and RLHF \cite{christiano2017rlhf} align reasoning with human preferences. In multimodal settings, models like LLaVA-CoT \cite{liu2023llava} and Multimodal-CoT \cite{zhou2023multimodal} demonstrate that reasoning traces enhance performance and interpretability. However, these advances remain underexplored in remote sensing, where reasoning over spatial relationships, multispectral features, and temporal change is essential.


Our work addresses this gap by unifying two complementary directions in remote sensing VLM research: reasoning-augmented supervision and preference-based alignment. Bringing these together in the geospatial context not only enables more interpretable and trustworthy models but also establishes a foundation for next-generation climate intelligence systems capable of supporting decision-making in high-stakes, complex scenarios.

\section{Datasets}
We conduct experiments on \textbf{RSVQA} (Low- and High-Resolution) and \textbf{FloodNet}, \ref{fig:datasets_examples}
with dataset-specific question types and answer formats summarized in Table~\ref{tab:datasets}. 
\textbf{RSVQA-LR} is built from \textit{Sentinel-2} imagery over the Netherlands at 10~m resolution~\cite{Lobry_2020}, 
while \textbf{RSVQA-HR} uses 15~cm aerial RGB images from the \textit{USGS HRO} collection covering U.S. urban areas. 
\textbf{FloodNet} contains high-resolution UAV imagery collected with \textit{DJI Mavic Pro} quadcopters during the 
Hurricane Harvey response (Texas/Louisiana, 2017), offering unique fidelity as real disaster-response imagery~\cite{floodnet}.

\section{Methodology}
Our pipeline consists of three main stages: (A) \textbf{CoT Data Distillation}, (B) \textbf{Supervised Fine-Tuning (SFT)}, and (C) \textbf{Reinforcement Learning with DPO}. Each stage builds on the previous one, progressively improving the model’s ability to reason over geospatial imagery while producing reliable and interpretable responses.
\begin{figure}[h!] 
    \centering
    \includegraphics[width=1\textwidth]{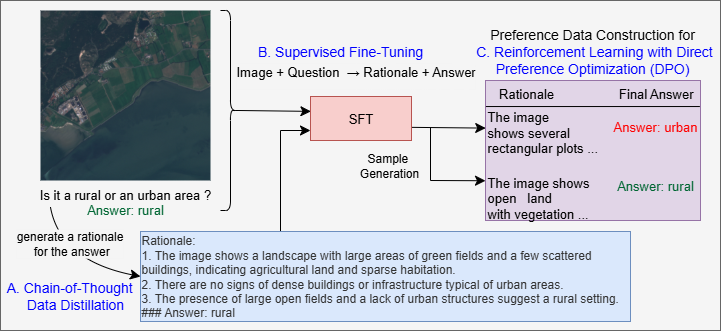} 
    \ \caption{Workflow diagram showing the methodology: 
    (A) Chain-of-Thought Data Distillation, 
    (B) SFT, 
    (C) Preference Data Construction for Reinforcement Learning with DPO.}
    \label{fig:sample}
\end{figure}

\paragraph{A. Chain-of-Thought Data Distillation}
The limited availability of \textbf{CoT-annotated data} for geospatial VQA poses a significant challenge. To address this, we employ a data distillation strategy that leverages existing direct answer data while enriching it with rationales. Specifically, the \textit{InternVL2-LLaMA3-76B} model ~\cite{chen2024internvlscalingvisionfoundation} is prompted with the image, question, and ground-truth answer to generate step-by-step reasoning \ref{gen} that leads to the correct prediction. This process allows us to construct synthetic rationale-augmented training data from otherwise rationale-free supervision.

To ensure quality, we introduce a verification step using \textit{Qwen2-VL-72B-Instruct} ~\cite{wang2024qwen2vlenhancingvisionlanguagemodels} as an evaluator \ref{score}. This model scores the rationales according to both their quality—defined in terms of coherence, factual grounding, and logical consistency—and their completeness. Samples with low-scoring rationales or with rationales that yield an incorrect final answer are discarded. The result is a curated dataset of high-quality rationale-augmented examples suitable for downstream fine-tuning.

\paragraph{B. SFT}
For fine-tuning, we use \textit{LLaMA3-LLAVA-NeXT-8B}, initialized with \textit{Open LLaVA-NeXT} weights~\cite{liu2024llavanext}. The training corpus contains two types of supervision: direct question–answer pairs and question–rationale–answer triples. We experiment with two parameter-efficient strategies. In the first, only the projection layer is updated while the backbone and the vision tower remain frozen. In the second, we unfreeze all parameters, including those of the vision tower (\textit{CLIP-ViT-L/14@336px}~\cite{radford2021learningtransferablevisualmodels}), allowing for end-to-end optimization.

Different prompt templates are used depending on the supervision type \ref{fig:io}. When training on direct QA data, the model is instructed to \texttt{Answer the question with a short answer}. In contrast, when training on rationale-augmented data, the prompt is modified to \texttt{Generate a reason first and then output a short answer}, where the rationale is followed by the answer in the format \texttt{\#\#\# Answer: <final\_answer>}. The model is trained for two epochs in both settings, ensuring exposure to both direct and rationale-augmented supervision.

\paragraph{C. Reinforcement Learning with DPO}
While SFT enables the model to produce rationales, it does not guarantee that these rationales are consistent with user preferences or that they reliably guide the model to correct answers. To further improve alignment, we employ DPO~\cite{rafailov2023dpo}, a reinforcement learning method that directly optimizes the policy model by contrasting preferred and non-preferred outputs. DPO formulates training as a binary cross-entropy objective that compares the model’s likelihood of generating positive versus negative responses. In doing so, it increases the probability of producing coherent and correct rationales while discouraging poor reasoning.

To construct preference pairs, we use the SFT model itself as the policy generator. For each input, eight candidate responses are sampled: four with a decoding temperature of 0.2 and four with a temperature of 0.6, ensuring a balance between determinism and diversity. Each response is evaluated against the ground-truth answer \ref{tab:dpo_examples}, and data points lacking at least one correct and one incorrect response are removed. From the remaining pool, one correct rationale–answer pair and one incorrect rationale–answer pair are randomly selected, forming a positive–negative pair for DPO training. This strategy yields a dataset that captures the range of plausible model behaviors while providing a clear supervision signal.

Formally, the DPO training dataset is defined as
\[
\mathcal{D}_{\text{DPO}}=\{(I,x,y^{+},y^{-})\},
\]
where \(I\) denotes the input image, \(x\) the question, \(y^{+}\) the preferred (positive) rationale–answer, and \(y^{-}\) the non-preferred (negative) counterpart. The optimization objective is
\begin{equation}
\mathcal{L}_{\text{DPO}}(\pi_{\theta};\pi_{\text{ref}}) \;=\;
- \mathbb{E}_{(I,x,y^{+},y^{-}) \sim \mathcal{D}_{\text{DPO}}}
\left[
\log \sigma \!\left(
\beta \left[
\log \frac{\pi_{\theta}(y^{+}\,|\,x,I)}{\pi_{\text{ref}}(y^{+}\,|\,x,I)}
-
\log \frac{\pi_{\theta}(y^{-}\,|\,x,I)}{\pi_{\text{ref}}(y^{-}\,|\,x,I)}
\right]
\right)
\right],
\label{eq:dpo}
\end{equation}
where \(\pi_{\theta}\) is the policy model to be optimized, \(\pi_{\text{ref}}\) is the reference model initialized with SFT weights, \(\sigma\) is the logistic function, and \(\beta\) is a scaling hyperparameter set to \(0.1\) in our experiments. This formulation encourages the model to assign higher probability to preferred rationales and lower probability to less desirable ones, thereby aligning the reasoning process more closely with correctness and user expectations.

\section{Results}

Our experiments demonstrate the effectiveness of incorporating \textit{CoT} supervision into vision-language models for remote sensing question answering. Compared to the direct SFT baseline, CoT-based SFT on the projection layer improved overall accuracy by \textbf{18.19\%}, highlighting the benefits of explicit reasoning signals. Applying Direct Preference Optimization (DPO) on top of CoT data provided a further \textbf{5.67\%} improvement, confirming the role of preference alignment in refining reasoning quality. The most substantial gains came from fine-tuning the entire model (vision encoder, projection layer, and language model), which resulted in a \textbf{34.9\%} improvement over the initial zero-shot baseline, achieving an overall accuracy of \textbf{82.77\%}.

\begin{table*}[htbp]
\centering
\scriptsize
\label{tab:rsvqa_results}
\renewcommand{\arraystretch}{1.3}
\begin{tabular}{lccccccc}
    \toprule
    \textbf{Approach} & \textbf{Total Samples} & \textbf{Correct} & \textbf{Overall Accuracy} & \multicolumn{4}{c}{\textbf{Type-wise Accuracy}} \\
    \cmidrule(lr){5-8}
     & & & & \textbf{Comp} & \textbf{Count} & \textbf{Presence} & \textbf{Rural/Urban} \\
    \midrule
    Initial Zero-Shot & 14339 & 6858 & 0.4783 & 0.3987 & 0.1757 & 0.6091 & 0.6667 \\
    SFT with Direct Data & 14339 & 6940 & 0.4840 & 0.4397 & 0.2619 & 0.5611 & 0.8333 \\
    SFT with CoT Data\\ (Projection Layer Only) & 14339 & 9548 & 0.6659 & 0.6881 & 0.1178 & 0.6890 & 0.7222 \\
    DPO on SFT CoT Data\\(Projection Layer Only) & 14339 & 10362 & 0.7226 & 0.7099 & 0.1652 & 0.7915 & 0.8333 \\
    SFT with CoT Data\\(All Weights Unfrozen) & 14339 & 11868 & \textbf{0.8277} & \textbf{0.8532} & \textbf{0.2337} & \textbf{0.8511} & \textbf{0.7778} \\
  
    \bottomrule
\end{tabular}
\caption{Performance comparison across fine-tuning strategies on RSVQA-LR and RSVQA-HR datasets. Results are reported as overall and type-wise accuracy for different question categories.}

\end{table*}

Our best-performing model also generates interpretable reasoning traces, which enhances user trust and improves transparency. This is particularly valuable in remote sensing applications, where understanding the rationale behind a prediction is as important as the prediction itself.

Despite these improvements, the model continues to struggle on counting-based questions, with type-wise accuracy lagging behind other categories. We hypothesize that this is due to limitations in text-based CoT explanations, which may not adequately represent precise numerical reasoning. Addressing this limitation could require integrating explicit counting modules or grounding mechanisms that align object-level detections with reasoning steps.

All models were trained on the \textbf{RSVQA-LR} and \textbf{RSVQA-HR} subsets and evaluated on the same benchmarks. We also evaluated our fine-tuned models on the \textbf{FloodNet} dataset. The model fine-tuned on direct data achieved an accuracy of \textbf{59.1\%}, while the model fine-tuned on CoT data achieved a higher accuracy of \textbf{67.4\%}, suggesting improved transferability with rationale-based training but also highlighting challenges in adapting to disaster-specific imagery. Training convergence was efficient, with the model stabilizing within two epochs.

\section{Conclusion}

This work shows that chain-of-thought supervision enhances geospatial VQA by improving both accuracy and interpretability. On RSVQA, CoT supervision yielded a 34.9\% accuracy gain over direct baselines, with DPO providing further refinement. On FloodNet, CoT also improved transfer performance (59.1\% → 67.4\%), highlighting stronger generalization to disaster imagery. Beyond accuracy, reasoning traces promote transparency and trust, which are vital in climate and disaster response. Remaining challenges, especially in counting and cross-dataset adaptation, suggest the need for explicit numerical reasoning and stronger transfer methods. Overall, structured reasoning emerges as a promising pathway toward reliable and trustworthy geospatial AI.

\bibliographystyle{unsrtnat}
\bibliography{refs}

\appendix
\section{Appendix}

\subsection{Datasets: Distribution and Visual Examples}
\begin{table}[h]
\centering
\begin{tabular}{l c l l}
\hline
\textbf{Dataset} & \textbf{Total Samples} & \textbf{Question Types} & \textbf{Answer Format} \\
\hline
\multirow{3}{*}{FloodNet} & \multirow{3}{*}{4510} 
& Simple/Complex Counting & Numerical \\
& & Condition Recognition & Flooded / Non-Flooded \\
& & Yes-No & Yes / No \\
\hline
\multirow{2}{*}{RSVQA-HR (15 cm)} & \multirow{2}{*}{62554} 
& Comparison (Comp) & Yes / No \\
& & Presence & Yes / No \\
\hline
\multirow{4}{*}{RSVQA-LR (10 m)} & \multirow{4}{*}{10004} 
& Comparison (Comp) & Yes / No \\
& & Presence & Yes / No \\
& & Counting & Numerical \\
& & Rural/Urban & Rural / Urban \\
\hline
\end{tabular}
\caption{Overview of datasets used in this work, including total number of samples, supported question types, and corresponding answer formats.}

\label{tab:datasets}
\end{table}
\begin{figure}[!htbp]
    \centering
    \begin{minipage}{0.30\textwidth}
        \centering
        \includegraphics[width=110pt]{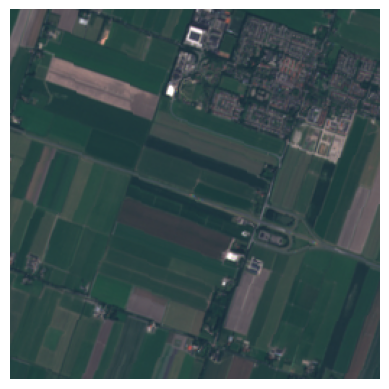}
        \caption{RSVQA-LR}
    \end{minipage}
    \hfill
    \begin{minipage}{0.30\textwidth}
        \centering
        \includegraphics[width=110pt]{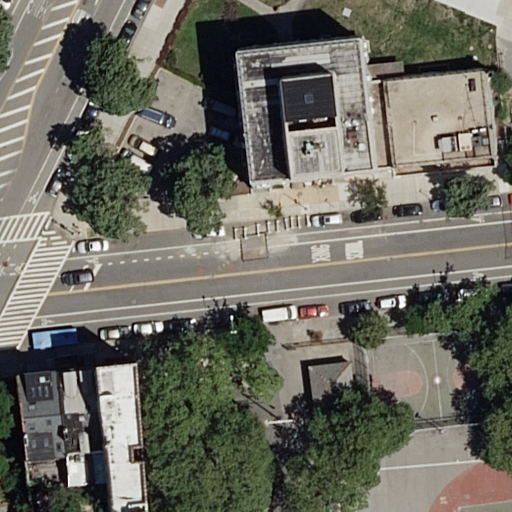}
        \caption{RSVQA-HR}
    \end{minipage}
    \hfill
    \begin{minipage}{0.38\textwidth}
        \centering
        \includegraphics[width=140pt]{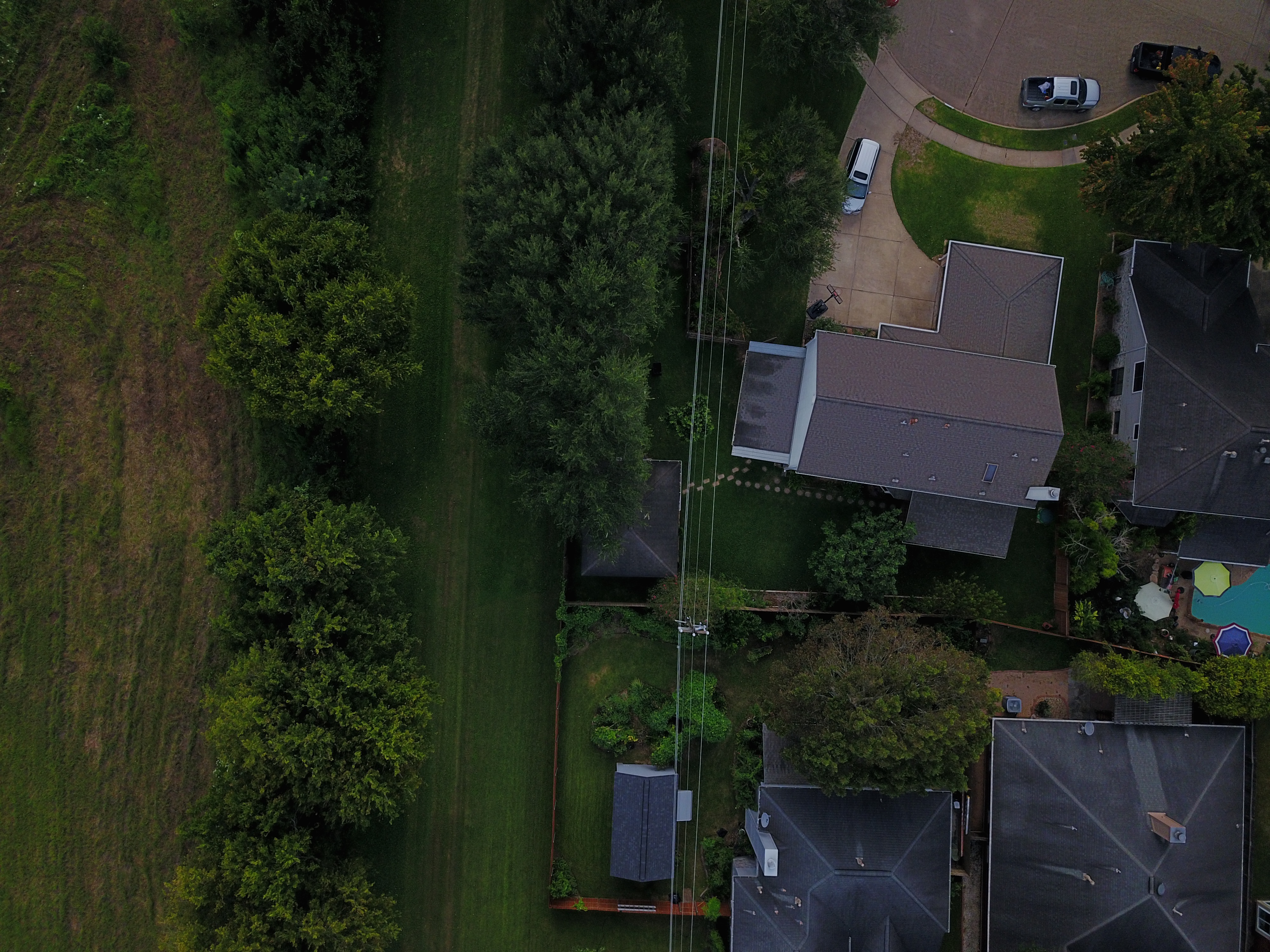}
        \caption{FloodNet}
    \end{minipage}
    \label{fig:datasets_examples}
\end{figure}

\subsection{Prompt Template for CoT Generation}

\begin{tcolorbox}[colback=pink!10!white, colframe=pink!50!black, title=Prompt for Rationale Generation,label={gen}]
\begin{verbatim}
<|image|> 
You are given a satellite image and a question, along with a
reference answer. Your task is to generate a detailed rationale 
that uses the visual content of the image to derive the answer.

## Question
{question}

## Reference Answer
{answer}

## Instructions
- Do not state an answer at the beginning
- Use detailed reasoning grounded directly in what you see in the
  image.
- Justify your conclusion clearly with observations.
- The reasoning should be clear, grounded in the visual content, 
  and should not speculate or estimate.
- Do not comment on the reference answer. Do not state whether the
  reference answer is correct or consistent to your finding.
  You are writing your own solution.
- Finish with a final derived one word answer on a new line as:


### Answer: <your_answer>
\end{verbatim}
\end{tcolorbox}
\subsection{Input-Output Example for Model Evaluation}
\begin{figure}[h!] 
    \centering
    \includegraphics[width=1\textwidth]{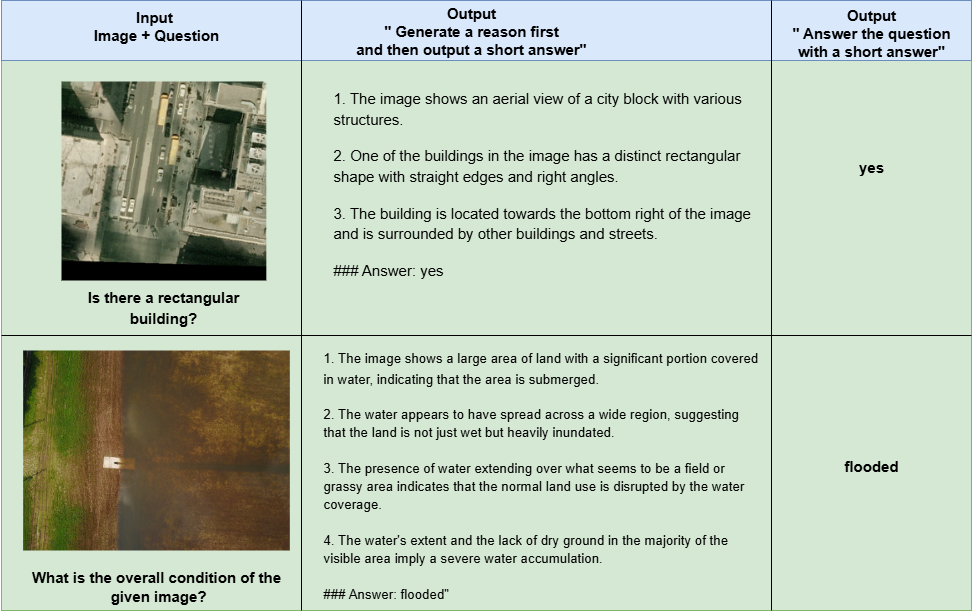} 
    \ \caption{Prompt-based Model Input and Output}
    \label{fig:io}
\end{figure}
\subsection{Prompt Template for Rationale Evaluation}
\begin{tcolorbox}[colback=blue!5!white, colframe=blue!50!black, title=Prompt for Rationale Scoring,label={score}]
\begin{verbatim}
You are a visual reasoning expert. Your task is to verify if a given
rationale (chain-of-thought) explanation is logically coherent based on 
the provided image, question, and answer.

Please follow these steps:
1. Read the question and analyze the image carefully.
2. Evaluate the rationale for:
   - Logical consistency
   - Visual grounding in the image
   - Relevance and coherence
   - Completeness (did it miss anything important?)
3. Determine whether the answer derived from the rationale follows
   naturally and reasonably and is valid based on the image and question.
4. Compare the provided answer with the ground truth answer.
5. If the answers differ, decide which answer is more likely
   correct.
6. Provide a quality score (0 to 10) indicating how confident you
   are in the rationale.
7. Poor structure, vague reasoning, or lack of detail should lower 
   the score.

Format as JSON:
  "rationale_logically_valid": true/false,
  "rationale_consistent_with_image": true/false,
  "rationale_correctness_score": float,
  "rationale_quality_score": float,
  "answer_follows_from_rationale": true/false,
  "answer_matches_ground_truth": true/false,
  "correct_answer": "model"/"ground_truth",
  "explanation": "Short explanation"
\end{verbatim}
\end{tcolorbox}

\subsection{Example: DPO Training Data Construction}

\noindent\textbf{Image + Question:}  
\begin{figure}[h!] 
    \centering
    \includegraphics[width=0.4\textwidth]{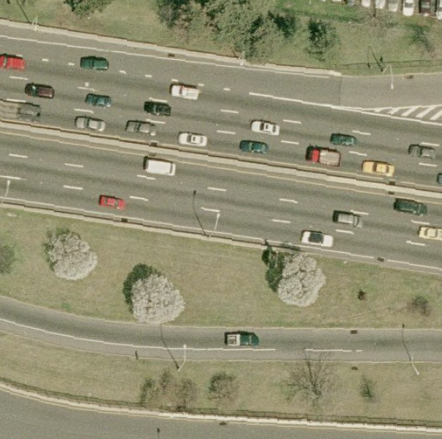} 
    \label{fig:ex_img}
\end{figure}

\textit{Are there more grass areas than roads? Answer in lower case. Generate a reason first and then output a short answer.}

\noindent\textbf{Ground Truth Rationale + Answer:} 

The image shows a large highway with multiple lanes of traffic, indicating a significant amount of road space. The grass areas are limited to small patches on the sides and median of the highway, which are much smaller in comparison to the road area. The majority of the image is dominated by the road, with vehicles and lanes clearly visible, while the grassy areas are minimal and scattered.

\texttt{\#\#\# Answer: no}

\begin{table}[h]
\centering
\renewcommand{\arraystretch}{1.3} 
\begin{tabular}{|p{0.9\textwidth}|}
\hline
The image shows a road with multiple lanes and a grassy area on the side. The road occupies a significant portion of the image, while the grassy area is relatively smaller. The road is clearly the dominant feature, with multiple lanes and vehicles, indicating it is the primary focus. The grassy area is limited to the side of the road and does not cover a large portion of the image. 
\textbf{\textcolor{green}{\#\#\# answer: no}} \\ \hline

The image shows a road with multiple lanes and a grassy area on the side. The road occupies a larger portion of the image compared to the grassy area. The grassy area is limited to the sides of the road, while the road stretches across the image, indicating a greater area of road compared to grass. 
\textbf{\textcolor{green}{\#\#\# answer: no}} \\ \hline

The image shows a large road with multiple lanes and a few grass areas. The road occupies a significant portion of the image, while the grass areas are relatively small. The road is clearly more extensive than the grass areas. 
\textbf{\textcolor{green}{\#\#\# answer: no}} \\ \hline

The image shows a highway with multiple lanes and a grassy area alongside it. There are more visible lanes than grassy areas. 
\textbf{\textcolor{green}{\#\#\# answer: no}} \\ \hline

The image shows a large road with multiple lanes and several vehicles. The surrounding area has a significant amount of grass, including a large green space and smaller patches of grass. The road occupies a significant portion of the image, while the grassy areas are more prominent and cover a larger area. 
\textbf{\textcolor{green}{\#\#\# answer: no}} \\ \hline

The image shows a highway with multiple lanes, and there are several grass areas visible on both sides of the road. The grass areas are more extensive than the road itself. 
\textbf{\textcolor{red}{\#\#\# answer: yes}} \\ \hline

The image shows a highway with multiple lanes and several vehicles. To the left of the highway, there is a grassy area with trees. The grassy area is significantly larger than the road lanes. The grass area occupies more space than the roads. 
\textbf{\textcolor{red}{\#\#\# answer: yes}} \\ \hline

The image shows a large area of grass on the left side and a long road on the right side. The grass area is significantly larger than the road area. 
\textbf{\textcolor{red}{\#\#\# answer: yes}} \\ \hline
\end{tabular}

\caption{Example of eight candidate responses for DPO preference data construction. Final answers are highlighted in \textcolor{green}{green (no)} and \textcolor{red}{red (yes)}.}
\label{tab:dpo_examples}
\end{table}


\end{document}